\documentclass[default]{sn-jnl}

\usepackage{graphicx}%
\usepackage{multirow}%
\usepackage{amsmath,amssymb,amsfonts}%
\usepackage{amsthm}%
\usepackage{mathrsfs}%
\usepackage[title]{appendix}%
\usepackage[table]{xcolor}
\usepackage{xcolor}%
\usepackage{textcomp}%
\usepackage{manyfoot}%
\usepackage{booktabs}%
\usepackage{algorithm}%
\usepackage{algorithmicx}%
\usepackage{algpseudocode}%
\usepackage{listings}%
\usepackage{array,tabularx}%
\usepackage{enumerate}
\usepackage{framed,enumitem} %
\usepackage{mathbbol}
\usepackage{url}
\usepackage[most]{tcolorbox}
\usepackage{hhline}
\usepackage{geometry}

\usepackage[T1]{fontenc}
\usepackage{fancyvrb,fvextra}
\usepackage[noframe]{showframe}
\usepackage{framed}
\usepackage{lipsum}
\usepackage[obeyspaces]{xurl}
 {\endMakeFramed}
\definecolor{shadecolor}{gray}{0.75}

\theoremstyle{thmstyleone}%
\theoremstyle{thmstyletwo}%
\theoremstyle{thmstylethree}%
\newcommand{\vect}[1]{\boldsymbol{#1}}

\newenvironment{conditions*}
  {\par\vspace{\abovedisplayskip}\noindent
   \tabularx{\columnwidth}{>{$}l<{$} @{}>{${}}c<{{}$}@{} >{\raggedright\arraybackslash}X}}
  {\endtabularx\par\vspace{\belowdisplayskip}}

\usepackage{verbatim}

\begin{document}

\title[Article Title]{Segmented Harmonic Loss: Handling Class-Imbalanced Multi-Label Clinical Data for Medical Coding with Large Language Models}


\author*[]{\fnm{Surjya} \sur{Ray}  \dgr{PhD}}
\author[]{\fnm{Pratik} \sur{Mehta}}
\author[]{\fnm{Hongen} \sur{Zhang} \dgr{MD}}
\author[]{\fnm{Ada} \sur{Chaman} }
\author[]{\fnm{Jian} \sur{Wang}  \dgr{PhD}}
\author[]{\fnm{Chung-Jen} \sur{Ho}  \dgr{PhD}}
\author[]{\fnm{Michael} \sur{Chiou} }
\author[]{\fnm{Tashfeen} \sur{Suleman} }      
      
\affil{CloudMedx Inc}


\abstract{The precipitous rise and adoption of Large Language Models (LLMs) have shattered expectations with the fastest adoption rate of any consumer-facing technology in history.  Healthcare, a field that traditionally uses NLP techniques, was bound to be affected by this meteoric rise.  In this paper, we gauge the extent of the impact by evaluating the performance of LLMs for the task of medical coding on real-life noisy data.  We conducted several experiments on MIMIC III and IV datasets with encoder-based LLMs, such as BERT.  Furthermore, we developed Segmented Harmonic Loss, a new loss function to address the extreme class imbalance that we found to prevail in most medical data in a multi-label scenario by segmenting and decoupling co-occurring classes of the dataset with a new segmentation algorithm.  We also devised a technique based on embedding similarity to tackle noisy data. Our experimental results show that when trained with the proposed loss, the LLMs achieve significant performance gains even on noisy long-tailed datasets, outperforming the F1-score of the state-of-the-art by over ten percentage points.}




\maketitle

\section{Introduction}\label{sec1}

In the fall of 2019, Google incorporated BERT \cite{bert} in its US search engine for English language queries \cite{bert-search}.  Two months later, they expanded the service to include over 70 languages worldwide for their non-English search engines \cite{bert-tweet}.  BERT, a Large Language Model (LLM), was developed and introduced by Google less than a year prior \cite{bert-blog}.  This upgrade of Google's search engine took place just two years after the introduction of BERT's spiritual parent, the game-changing transformer architecture, by Vaswani et al. in 2017 \cite{attention}.  Based on the encoder part of the original transformer architecture, BERT is excellent at Natural Language Understanding.  This capability was essential to interpret and understand these unknown queries that the search engine faced, which accounts for 15\% of its whopping 8.5 billion daily queries \cite{bert-search}.  It was a gutsy move demonstrating its faith in the nascent technology, considering its `search and other' revenues accounted for 58\% of Alphabet's total revenue \cite{search-revenue}.  

Google was far from the only company aiding this meteoric rise and adoption of LLMs.  ChatGPT, OpenAI's transformer-based chatbot, garnered 100 million active users in January 2023, just two months after launch.  It is the fastest adoption rate of any consumer-facing technology in history \cite{chatgpt-adpotion}.  Healthcare, which has a long history of repurposing Natural Language Processing (NLP) techniques to interpret mammoth volumes of Electronic Medical Records (EMRs) and clinical notes, was bound to be affected by this LLM wave.  The more pertinent question was what this adoption would look like and its extent.  

Google was already investigating the possibilities of LLMs in healthcare with Med-PaLM \cite{med-palm}, based on their Pathways Language Model, PaLM, a 540-billion parameter LLM \cite{palm}.  At the same time, the World Health Organization called for caution against this precipitous adoption of LLMs for health-related purposes \cite{who-llm}.  Without checks and balances, the decoder-based LLMs are prone to hallucinations, where a response may appear correct and relevant to those untrained in healthcare but is, in fact, erroneous.  It may lead to ``errors by healthcare workers, cause harm to patients, erode trust in AI, and thereby undermine or delay the potential long-term benefits and uses of such technologies around the world,'' they worried \cite{who-llm}.  

To this end, we gauge the extent of the long-term usefulness of LLMs in healthcare, a field that requires stringent quality control due to strict regulations.  It is a multifaceted initiative requiring experimentation on real-life healthcare data on various healthcare use cases.  This paper reflects our initial exercises in that investigation.  To make such a broad scope manageable, we picked the use case of medical coding of clinical notes for our initial assessment.  Medical coding, which is the process of assigning standardized codes to a patient's medical information, is a segment projected to reach $\$38$ billion by 2030.  AI-aided medical coding is supposed to take at least $\$5.71$ Billion out of it \cite{medical-coding-market}.  This paper focuses on ICD codes (version 9 or ICD-9) \cite{icd9}, a system designed by WHO and used by healthcare professionals to classify and code diagnoses and procedures for claims processing.  However, the methodology and implications that came out of this paper apply to all kinds of clinical coding.

As with all data-centric ML paradigms, we started with an in-depth analysis of various real-life datasets of unstructured EMRs, specifically MIMIC III and IV \cite{mimiciv_v1}.  We quickly realized this data is highly imbalanced and heavily reliant on the note-taking practices of healthcare providers.  These findings guided our research to realize the full potential of LLM for medical coding.  Our key contributions can be summarized as follows: (1) We used a novel technique utilizing embedding similarity to preprocess and create labeled data from the two MIMIC datasets.  Since the input of BERT is limited to 512 tokens, we could only include a fraction of the clinical notes that serve as inputs to our models.  The method was instrumental in removing those ICD-9 codes that were absent in the chunk of input text, thereby reducing false positives in the training set.  (2) We developed a new loss function, Segmented Harmonic Loss, to deal with the extreme class imbalance in a multi-label scenario after experimentation with existing methods, such as Focal Loss \cite{focal} or Class Balanced Loss \cite{CBLoss}, which failed to mitigate the problem effectively.  We believe our study will provide practical guidelines for dealing with medical or other imbalanced data to unlock the full potential of data-centric ML approaches to harness the power of LLMs.  (3) We achieved state-of-the-art MIMIC dataset benchmarks on medical coding, besting the previous results by more than ten percentage points, showcasing the potential of domain-specific LLMs in healthcare.

\section{Data}\label{sectionData}
Domain-specific finetuning of LLMs requires a considerable amount of data.  In healthcare, Patient Health Information (PHI) is highly regulated by various laws to protect patient privacy, which varies from country to country.  In the U.S., the Health Insurance Portability and Accountability Act of 1996 (HIPAA) \cite{hippa} regulates how PHI may be released.  The law specifies 18 types of identifiers, such as name, address, telephone numbers, medical record number, and other fields that must be deidentified before releasing any PHI.  As a result, reliable and standardized open-source PHI and EMRs for research are hard to come by.  The two data sources that meet the reliability and quality control required for research are MIMIC III and MIMIC IV datasets \cite{mimiciv_v1}.  To date, these are the only large-scale, reliable, and freely accessible electronic health record datasets available \cite{mimic-nature} from the U.S.  MIMIC III has around $60 K$ records of PHI.  MIMIC IV has around $300 K$, of which $200 K$ are ICD-9 codes.  The remaining $100 K$ are the newer ICD-10 codes.  In order to maximize our data, we created an amalgamated dataset by extracting and combining the ICD-9 records from these two datasets.  This process of preparing and preprocessing this dataset is described in the following subsections. 

\subsection{Data Format and Structure} \label{d1}

Our data is comprised of two components: clinical notes ($\vect{x}$) and a list of corresponding codes assigned to the note ($\vect{y}$). Notes are unstructured text containing healthcare information like symptoms, diagnosis, medical services, and procedures noted by physicians and healthcare providers.  Codes are the corresponding ICD-9 codes designated to the note by clinical coders.  Together, they form one record of a note-code pair $(\vect{x}, \vect{y})$. Denoting the $k^{th}$ note-code pair as $(\vect{x}^{k}, \vect{y}^{k})$, we can define our dataset as $\mathcal{D} = \{(\vect{x}^{1}, \vect{y}^{1}), (\vect{x}^{2}, \vect{y}^{2}), ..., (\vect{x}^{N}, \vect{y}^{N}) \}$ where $N$ is the number of training samples and $k \in \{1,2, ..., N \}$.  For the scope of our problem, we consider $\mathcal{C}$ codes or classes. Since we can have multiple ICD-9 codes (diagnoses, procedures and so on) for each note, the label for the $k^{th}$ sample, $\vect{y}^{k}$, can be written as:

\begin{equation}
	\vect{y}^{k} =[y^{k}_{1},y^{k}_{2}, ..., y^{k}_{\mathcal{C}} ]^T  \label{eq:yk_1}
\end{equation}
where:\\
\begin{conditions*}
 y^{k}_{i}     	& :		& label for the $i^{th}$ class/code, \&  $y^{k}_{i} \in\{0, 1\},$\\
 i 				& \in   & $\{ 1, 2, ..., \mathcal{C}\}$.      
\end{conditions*}

An example of note-code pair before vectorization is show in the Listing \ref{example_x} and Table \ref{taby}. 

\begin{lstlisting}[label=example_x,caption=A truncated example of a note ($\vect{x}$) from the combined MIMIC dataset,frame=tb][h]

symptomatic bradycardia permanent pacemaker placement chronic systolic 
congestive heart failure coronary artery disease lastname 66794**] is 
an -- year old gentleman with a history of atrial fibrillation on 
amiodarone and coumadin cad with 3vd seen on cardiac cath in [**2186**]
chf with ef of 20\% who initially presented to [**hospital1 **] 
[**location (un) 620**] last night complaining of a slow heart beat. 
per his wife at around midnight last night he came to bed complaining 
that his heart was beating very slowly he felt weak and that he could 
feel it in his chest. at that time he asked his wife to call 911 and 
he was taken to [**hospital1 **] [**location (un) 620**] for further 
evaluation. per his wife he did not complain of any chest pain 
shortness of breath or nausea. also the family notes that he has been 
having some falls/syncope at home including one event that he did not 
remember in the past few weeks. . per ems report he was found to be 
pale cool and diaphoretic on initial examination. ...
\end{lstlisting}

\begin{table}[h]
\caption{Codes ($\vect{y}$) and their descriptions}\label{taby}%
\centering
\begin{tabular}{ | c | c | }
  \hline
  Codes ($y_{i}$) 	& Description 								\\ \hline            
  $42789$ 			& Other specified cardiac dysrhythmias 		\\ \hline  
  $4263$ 			& Other left bundle branch block 	       	\\ \hline  
  $2449$ 			& Acquired hypothyroidism 	                	\\ \hline  
  $42822$ 			& Chronic systolic heart failure 	        \\ \hline  
  $41401$ 			& Coronary atherosclerosis of artery 	    \\ \hline  
  $3659$ 			& Glaucoma 	                					\\ \hline  
  $4280$ 			& Congestive heart failure 	               	\\ \hline       
  
\end{tabular}
\end{table}

\subsection{Frequency Thresholding: Removal of Rare Codes} \label{d2}

\begin{figure*}
\centering
\includegraphics[width=1.0\textwidth]{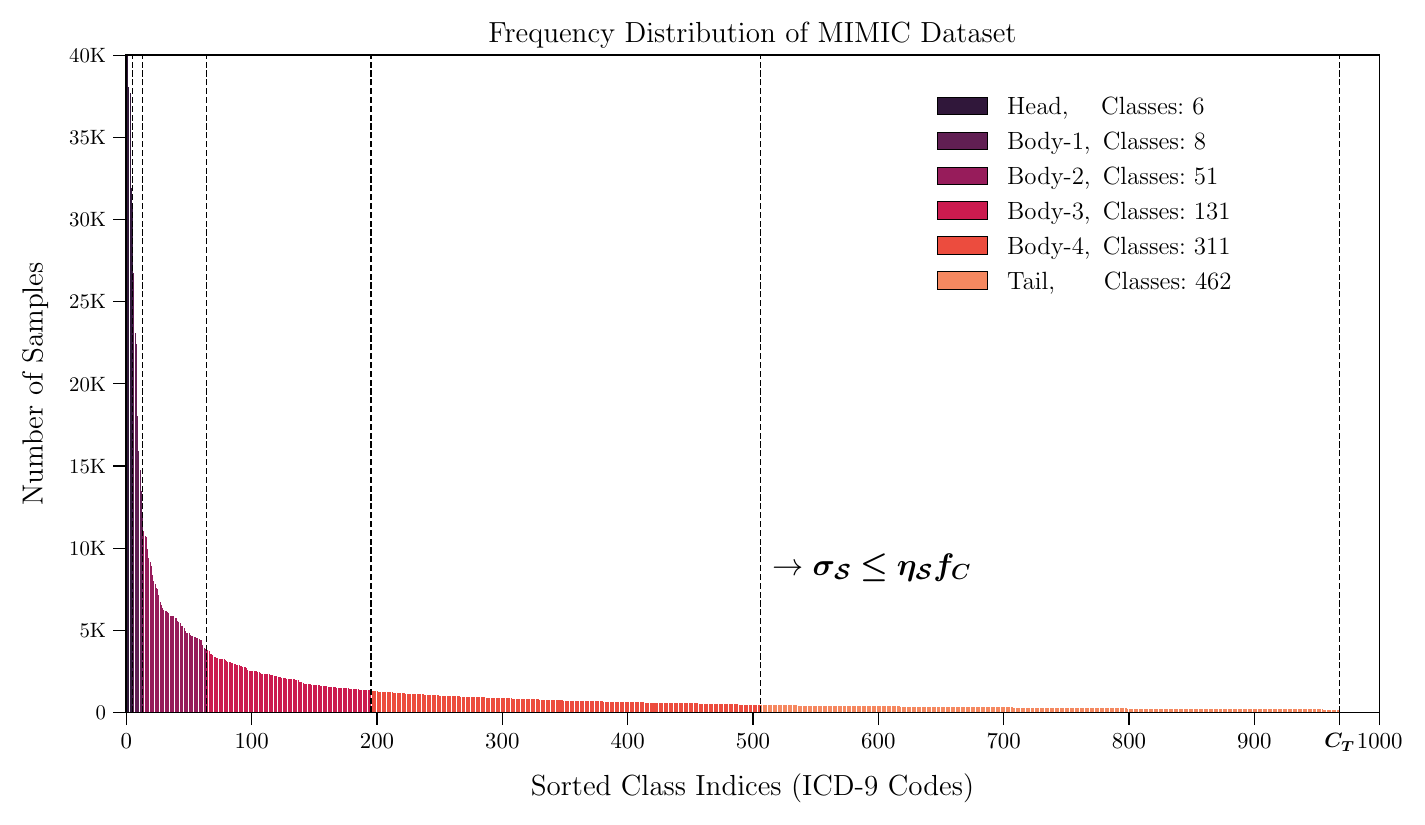}
\caption{The frequency distribution of the classes (ICD-9 codes) of the MIMIC dataset follows a long-tailed distribution. We use a recursive algorithm to segment the classes such that the standard deviation, $\sigma_{\mathcal{S}}$, of the tail segment $\mathcal{S}$, does not exceed a fraction of the frequency, $f_{\mathcal{C}}$, of the last class $\mathcal{C}$. The fraction, $\eta_{\mathcal{S}}$, is a hyperparameter set to 0.5 in our experiments. Following this recursive algorithm, the classes of our dataset are divided into 6 segments with a head region, a tail region and four intermediate `body' regions. \textbf{NOTE:} The frequency axis is shown upto 40K. The head classes actually goes upto around 80K.}\label{fig1}
\end{figure*}

The combined dataset generated $261953$ records or note-code pairs covering $9412$ unique ICD-9 codes.  The frequency distribution of the codes, sorted by frequency, is shown in Fig \ref{fig1}.  We excluded codes with frequencies less than 200 as, at such low frequencies, they were not enough to train an LLM.  As seen from Fig. \ref{fig1}, the frequency distribution of ICD-9 codes shows extreme class imbalance.  It follows a long-tailed distribution.  The `head' part of the distribution contains only a few ICD-9 codes that occur frequently.  The `tail' end of the distribution contains a large number of codes whose frequency is several orders of magnitude lower than those codes in the `head' section.  In fact, the frequencies of the head classes are $\approx~700$ times greater than those of the tail classes.  In the first step of data preprocessing, we enforced a frequency threshold of 200 to the code list of every note-code pairs.  This process removed 7924 unique codes from the code lists, leaving our dataset with 1488 codes with frequencies 200 and above.

\subsection{Creation of the Input Notes ($\vect{x}$)} \label{d3}
The raw clinical notes were large, sometimes pages long, and were primarily free-from.  With our medical professionals' aid, we found that the diagnoses were primarily present in three essential sections: 

\begin{enumerate}[label=\roman*.]
  \item \textit{Discharge Diagnosis} 
  \item \textit{History of Present Illness}
  \item \textit{Past Medical History}
\end{enumerate}

The last section was only relevant for diagnoses from past visits.  In this paper, we focus only on current diagnoses.  Hence, we created our inputs, $\vect{x}$, concentrating only on the first two sections, which are relevant for the current diagnoses.  We also replaced medical abbreviations with corresponding texts, improving our results.  In the provided example in \ref{example_x}, the abbreviation `CHF' was replaced with `Congestive Heart Failure.' If the text was longer than the input could accommodate, we truncated it to fit the input size of our LLM, which is limited to 512 tokens.  If the text was shorter than the input size, we padded it to the maximum size of 512 tokens. 

\subsection{Similarity Thresholding: Removal of Invalid Codes} \label{d3}

Because of the truncation step, many of the codes of label $\vect{y}$ are not represented by the input note, $\vect{x}$, i.e., the diagnoses of these codes  are not present in the truncated text for two reasons:

\begin{enumerate}[label=\roman*.]
  \item The text of $\vect{x}$ does not contain past medical history and hence does not contain codes from past visits.
  \item Some codes from the current visit may not be represented in the text $\vect{x}$, as the truncation had removed the text containing the diagnoses.
\end{enumerate}

\vspace{2pt}

\begin{figure*}[h]
\centering
\includegraphics[width=1.0\textwidth]{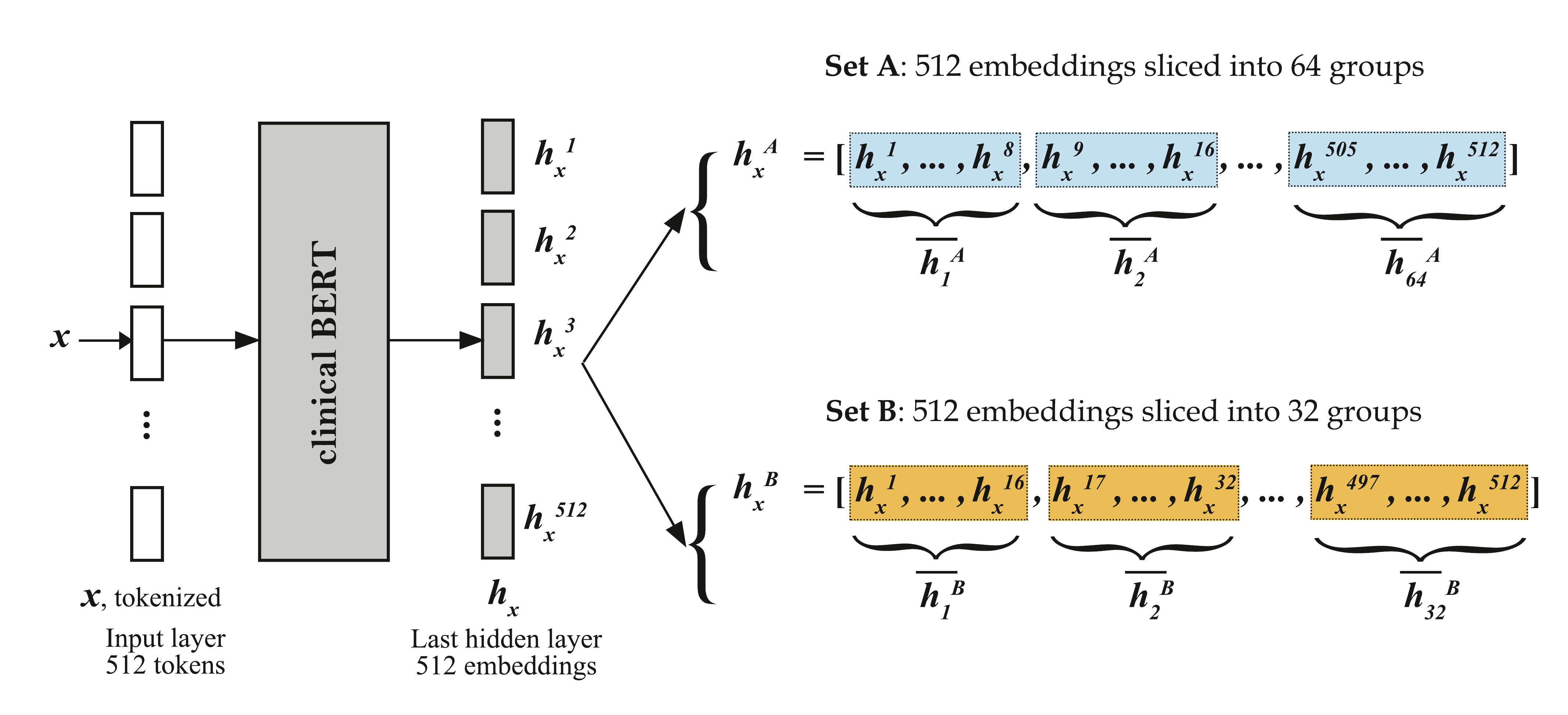}
\caption{The creation of two sets of mean embeddings using the output hidden state of ClinicalBERT}\label{fig_embed}
\end{figure*}

We needed to remove these invalid codes from $\vect{y}$ as these count as false positives.  To achieve this, we used a technique of comparing embeddings using cosine similarities.  To calculate the embeddings, we used the base model of clinicalBERT \cite{clinicalBERT}, which was trained on a large dataset with a large corpus of 1.2B words of diverse diseases created from EHRs from over 3 million patient records.  We found it to be very efficient in incorporating bidirectional medical context and clinical meaning compared to similar encoder models.  We passed each input note $\vect{x}$ through ClinicalBERT and saved the last hidden state, $\vect{h_{x}}$, as shown in Fig \ref{fig_embed}.  It is given by:

\begin{equation}
	\vect{h_{x}} =[\vect{h}^{1}_{\vect{x}},\vect{h}^{2}_{\vect{x}}, ..., \vect{h}^{512}_{\vect{x}} ]^T  \label{eq:hx}
\end{equation}
where $\vect{h}^{i}_{\vect{x}}$ is the $i^{th}$ embedding out of the 512 embeddings in the final hidden state.

We use $\vect{h_{x}}$ to create two sets of average embeddings, \textbf{Set A} and \textbf{Set B}.  As shown in Fig \ref{fig_embed}, we divide the 512 embeddings of \textbf{Set A} into 64 groups, each containing eight embeddings.  For each of these 64 groups, we calculate the mean embedding by averaging over the group.  As seen from Fig \ref{fig_embed}, this produces a set of 64 mean embeddings given by:

\begin{equation}
	\vect{h}^{A}_{\vect{x}} =[\bar{\vect{h}^{A}_{1}},\bar{\vect{h}^{A}_{2}}, ..., \bar{\vect{h}^{A}_{64}} ]^T  \label{eq:hA}
\end{equation}

where $\bar{\vect{h}^{A}_{i}}$ is the mean embedding of the $i^{th}$ group.

For \textbf{Set B}, we follow the same procedure as for \textbf{Set A}, except we divide the 512 embeddings into 32 groups of 16 embeddings. As for \textbf{Set A}, we compute the mean embeddings for these 32 groups giving:

\begin{equation}
	\vect{h}^{B}_{\vect{x}} =[\bar{\vect{h}^{B}_{1}},\bar{\vect{h}^{B}_{2}}, ..., \bar{\vect{h}^{B}_{32}} ]^T  \label{eq:hB}
\end{equation}

where $\bar{\vect{h}^{B}_{i}}$ is the mean embedding of the $i^{th}$ group.

We treat each ICD-9 code present in $\vect{y}$ separately.  For each $y_{i}$ in $\vect{y}$, we pass its official clinical text description (usually a few words) through ClinicalBERT.  For example, if $y_{i}$ corresponds to the code $42789$ whose textual description is `Other specified cardiac dysrhythmias,' we tokenize this text and pass it through the encoder.  At the output, we get an equal number of embeddings as the number of input tokens. i.e., if the input text description had $n$ tokens, the output hidden state, $\vect{h}_{y_{i}}$ will have $n$ embeddings.  We create a mean embedding, $\bar{\vect{h}_{y_{i}}}$ by summing over all of these $n$ embeddings.  

To find out if $y_{i}$ is present in the text in $\vect{x}$, we do the following:
\begin{enumerate}
  	\item Compute cosine similarities of $\bar{\vect{h}_{y_{i}}}$ with the 64 mean embeddings of $\vect{h}^{A}_{\vect{x}}$ in Equation \ref{eq:hA}.
  	\item Compute cosine similarities of $\bar{\vect{h}_{y_{i}}}$ with the 32 mean embeddings of $\vect{h}^{B}_{\vect{x}}$ in Equation \ref{eq:hB}.
	\item Take the maximum similarity score of these two sets as the representative similarity score of $y_{i}$ with $\vect{x}$.
\end{enumerate}

The textual descriptions of $y_{i}$ are usually a few words which results in a small number of input tokens $\ll$ the 512 input tokens from $\vect{x}$. Slicing the 512 output embeddings of $\vect{x}$ into chuncks of 8 and 16 allows focusing on different parts of the input text.  For smaller descriptions of $y_{i}$, usually, \textbf{Set A} performs better and for slightly larger descriptions of $y_{i}$, \textbf{Set B} performs better. We found that taking the maximum of the similarities of these two sets provided an optimal result with a similarity threshold of $0.55$. We also tried taking cosine similarity of $\bar{\vect{h}_{y_{i}}}$ with the mean of 512 embeddings of $\vect{x}$.  This does not perfom well as averaging over so many embeddings dilutes the values of the relevant embeddings.  

Table \ref{tab2} shows cosine similarity of the codes of $\vect{y}$ associated with note $\vect{x}$ in Listing \ref{example_x}. The correct codes are chosen as those
with a similarity above $0.55$ . On inspection of the note $\vect{x}$, the selection of codes 42789, 42822, 41401 and 4280 are found to be correct. 

\begin{table}[h] 
\caption{Codes ($y_{i}$) and their Similarity Scores with $\vect{x}$}\label{tab2}%
\centering
\begin{tabular}{ | c | c | c | }
  \hline
  Codes ($y_{i}$) 	& Description 					& Similarity Score \\ \hline            
  $42789$ 			& Other specified cardiac dysrhythmias 			& $\textbf{0.64}$ \\ \hline  
  $4263$ 			& Other left bundle branch block 	        		& $0.50$ \\ \hline  
  $2449$ 			& Acquired hypothyroidism 	                		& $0.53$ \\ \hline  
  $42822$ 			& Chronic systolic heart failure 	           	& $\textbf{0.72}$ \\ \hline  
  $41401$ 			& Coronary atherosclerosis of artery 	       	& $\textbf{0.66}$ \\ \hline  
  $3659$ 			& Glaucoma 	                						& $0.44$ \\ \hline  
  $4280$ 			& Congestive heart failure 	               		& $\textbf{0.72}$ \\ \hline       
  
\end{tabular}
\end{table}

\vspace{2pt}

After the step of similarity thresholding, codes that are not represented in the input text are removed form $\vect{y}$. This alters the frequency distribution of the codes once more. Hence, we do frequency thresholding on the codes as before, removing any code that occurs less than 200 times in the dataset. After this
exercise, we are left with 255736 records and a code set of 969 codes, which serves as our final dataset. 

\section{Segmented Training} \label{algorithm}

We tried loss functions designed to mitigate class imbalance in multi-label scenarios, which we have included in our results in Section \ref{experiments}. With enough hyperparameter-tuning, these specialized loss functions improved the $F_1$ scores and other metrics over the standard Binary Cross Entropy (BCE) loss function, but the performance nevertheless degraded for the less-represented classes towards the tail region of the distribution. We surmise this is because a singular model could not fit the $1000 \times$ order of discrepancy between the head and tail classes, especially with real-life noisy datasets. This led us to a different approach: fit separate models for different segments of the frequency distribution while decoupling co-occurring classes belonging to different segments using a new loss function. We formulate our problem mathematically in Subsection \ref{problem_formulation}, describe our segmentation algorithm in Subsection \ref{segmentation_algorithm}, and discuss the loss function in Subsection \ref{harmonic_loss}. 


\subsection{Problem Formulation}\label{problem_formulation}
As discussed in Subsection \ref{d1}, our data are comprised of note-code pairs. We denote the $k^{th}$ note-code pair as $(\vect{x}^{k}, \vect{y}^{k})$, and our training dataset as 
$\mathcal{D} = \{(\vect{x}^{1}, \vect{y}^{1}), (\vect{x}^{2}, \vect{y}^{2}), ..., (\vect{x}^{N}, \vect{y}^{N}) \}$ where $N$ is the number of training samples and $k \in \{1,2, ..., N \}$.  For the scope of our problem, we consider $\mathcal{C}$ codes or classes. We sort and index these classes according to their frequencies as shown in Fig \ref{fig1}. This sorted list is denoted by $\mathcal{F}$ where any index is a class id and the value at the index is the corresponding class frequency. This implies that two classes with indices $l$ and $m$ and frequencies $f_{l}$ and $f_{m}$, will follow the relation $f_{l} \geq f_{m}$ if $l < m$ and $l, m \in \{1,2, ..., \mathcal{C} \}$.  Since we can have multiple ICD-9 codes for each note, the label for the $k^{th}$ sample, $\vect{y}^{k}$, can be written as:

\begin{equation}
	\vect{y}^{k} =[y^{k}_{1},y^{k}_{2}, ..., y^{k}_{\mathcal{C}} ]^T  \label{eq:yk}
\end{equation}
where:\\
\begin{conditions*}
 y^{k}_{i}     	& :		& label for the $i^{th}$ class/code, \&  $y^{k}_{i} \in\{0, 1\},$\\
 i 				& \in   & $\{ 1, 2, ..., \mathcal{C}\}$.      
\end{conditions*}
 
This is a multi-label classification problem, i.e. the occurrence of $y^{k}_{i} = 1$ and $y^{k}_{j} = 1$, where $ i \neq j$ and $i, j \in \{1, 2, ...., \mathcal{C}\}$, are not necessarily mutually exclusive. Thus, given a clinical note $\vect{x}^{k}$, our objective is to predict the associated classes(ICD-9 codes), i.e. $\vect{y}^{k}$.

\subsection{Segmentation Algorithm} \label{segmentation_algorithm}
The Segmentation Algorithm divides the list $\mathcal{F}$ into $\mathcal{S}$ segments. The $r^{th}$ segment, $\mathcal{F}_{r}$, has $c_r$ classes such that 
$\sum_{r=1}^{\mathcal{S}} c_{r} = \mathcal{C}$.  The algorithm starts by creating the tail segment, $\mathcal{F}_{\mathcal{S}}$,  by grouping the last 
$c_{\mathcal{S}}$ classes of the tail such that there is a relatively small variation between their class frequencies. The amount of tolerable variation is 
controlled by a hyperparameter $\eta_{\mathcal{S}}$ such that:

\begin{equation}
	\sigma_{\mathcal{S}} \leq \eta_{\mathcal{S}}f_{\mathcal{C}}. \label{eq:sigma}
\end{equation}
where:\\
\begin{conditions*}
 f_{\mathcal{C}}     	& :		& the frequency of the last class $\mathcal{C}$.\\
 \sigma_{\mathcal{S}} 	& :		& the standard deviation of the class frequencies of segment $\mathcal{F}_{\mathcal{S}}$.\\
 \eta_{\mathcal{S}}		& :		& a hyperparameter in $(0, 1]$ which determines the tolerable variance, $\sigma^2_{\mathcal{S}}$, between the class frequencies 
 of segment $\mathcal{F}_{\mathcal{S}}$. \\
\end{conditions*}
For example, if $\eta_{\mathcal{S}}$ is set to 0.5, $\sigma_{\mathcal{S}}$ can be at most half of the last class frequency, $f_{\mathcal{C}}$. Thus, the choice 
of $\eta_{\mathcal{S}}$ shall control the amount of tolerable  variation between frequencies in a segment and will in turn control i) the number of segments, 
$\mathcal{S}$, and ii) the size of the segments i.e. the number of classes in these respective segments.

\begin{algorithm}
\caption{Partition list $\mathcal{F}$ into segments}
\begin{algorithmic}[1]
\State \text{//$segments$: List of segments.}
\State $segments \gets  [$  $]$ 
\State \text{//$\mathcal{F}$: index-sorted list of class frequencies} \\
\Procedure{SegmentF}{$\mathcal{F}$, $\eta_{\mathcal{S}}$, $segments$}
	
	\State $n \gets$ {$len(\mathcal{F})$}
	
	\State \text{// If list $\mathcal{F}$ is not empty, create a segment $\mathcal{F}_{curr}$ from the tail end }
	\If {n $>$ 0} 
	\State \text{// Get index of the last class of $\mathcal{F}_{curr}$ }
	\State $i_{end} \gets n$
	\State \text{// Get index of the first class of $\mathcal{F}_{curr}$ }
	\State $i_{start}\gets$ \Call{SegmentTail}{$\mathcal{F}$,  $\eta_{\mathcal{S}}$}
	\State \text{// Create the current segment $[i_{start}, i_{end}]$ }
	\State$\mathcal{F}_{curr}\gets$ $[i_{start} , i_{end}]$
	\State \text{// Add the current segment to list $segments$.}
	\State $segments.append(\mathcal{F}_{curr})$
	\State \text{// Update $\mathcal{F}$ by discarding segment $\mathcal{F}_{curr}$ }
	\State $\mathcal{F} \gets$ $\mathcal{F}[1:i_{start}]$
	\State \text{// Recursively call the procedure.}
	\State \Call{SegmentF}{$\mathcal{F}$, $\eta_{\mathcal{S}}$, $segments$}
	\EndIf
    \State \textbf{return}
\EndProcedure
\end{algorithmic}
\label{alg:segment}
\end{algorithm}

Once segment $\mathcal{F}_{\mathcal{S}}$ has been generated, we are left with $\mathcal{C^{'}} = \mathcal{C} - c_{\mathcal{S}} $ classes which 
approximate another long-tailed distribution. We recursively create more segments from the tail end, until there are no more 
classes left. The pseudo-code of the segmentation algorithm is shown in Algorithm \ref{alg:segment}. The procedure \textsc{SegmentF} 
recursively segments the list $\mathcal{F}$ from the tail end such that Equation \ref{eq:sigma} is satisfied. Inside this procedure, another 
procedure, \textsc{SegmentTail}, is used to create one segment at a time from the tail end. This procedure returns the index of the first class to be 
included in the segment being computed. It is denoted by $i_{start}$ in the pseudocode. The procedure \textsc{SegmentTail} maybe efficiently 
implemented using Binary Search and its pseudocode is presented in Appendix \ref{secA2}. Following this algorithm, our dataset is divided 
into six segments ($\mathcal{F}_{1}$ through  $\mathcal{F}_{6}$) as shown by the dotted vertical lines in Fig \ref{fig1} with $\mathcal{S} = 6$. 

Once $\mathcal{F}$ is segmented into $\mathcal{S}$ segments, the $k^{th}$ sample $\{\vect{x}^{k}, \vect{y}^{k}\}$, presented in Equation \ref{eq:yk} 
yields $\mathcal{S}$ labels $\vect{y}^{k,r}$, one for each segment, where the second superscript indicates the segment number, i.e., 
$r \in \{1, 2, ..., \mathcal{S}\}$. Thus, the label for the $k^{th}$ note-code pair $(\vect{x}^{k}, \vect{y}^{k,r})$ of segment $\mathcal{F}_{r}$ is:
\begin{equation}
	\vect{y}^{k,r} =[y^{k,r}_{\rho + 1},y^{k,r}_{\rho + 2}, ..., y^{k,r}_{\rho + c_{r}} ]^T  \label{eq:ykr}
\end{equation}

where:\\
\begin{conditions*}
 c_{r}				& :		& number of classes in segment $\mathcal{F}_{r}$, \\
\rho					& : 		& total number of classes in the first $r - 1$ segments = $\sum_{p=1}^{r-1} c_{p}$, \\
 y^{k, r}_{i}    & :		& label for the $i^{th}$ class/code \& $y^{k, r}_{i} \in \{0, 1\}$,\\
 i 					& \in   	& $\{ \rho + 1, \rho + 2, ...,\rho + c_{r}\}$.      
\end{conditions*}

Although we derive the segmentation algorithm using the example of a long-tailed imbalanced distribution, the algorithm works for all kinds of
imbalanced distributions as long as we order the classes according to their frequencies. 

We train each segment model on all $N$ training samples. This helps in decoupling co-occurring classes (ICD-9 codes). Consider two 
co-occurring classes $i$ and $j$ for the $k^{th}$ label, $\vect{y}^{k}$. This implies $y^{k}_{i} = 1$ and $y^{k}_{j} = 1$, where 
$ i \neq j$ and $i, j \in \{1, 2, ...., \mathcal{C}\}$. Let's assume $ i \in \mathcal{F}_{r}$ and $ j \notin \mathcal{F}_{r}$. When training for 
the segment $\mathcal{F}_{r}$, the sample $\vect{y}^{k,r}$, derived from $\vect{y}^{k}$, shall ignore the class $j$, thereby teaching the 
model of segment $\mathcal{F}_{r}$ to discriminate against the $j^{th}$ class. This decouples the class $j$ from segment $\mathcal{F}_{r}$.

If $\vect{x}^{k}$ does not represent any classes for segment $\mathcal{F}_{r}$, its label, $\vect{y}^{k,r}$, will be a zero vector i.e. 
$\vect{y}^{k,r} = [0, 0, ..., 0]^T$. These serve as negative samples for segment $\mathcal{F}_{r}$. The number of negative samples 
contributed by segments other than $\mathcal{F}_{r}$ may be orders of magnitude more than positive samples coming from $\mathcal{F}_{r}$. 
Unless suitably down-weighted, the loss contributions from these negative samples will overwhelm the training loss while using a 
traditional loss function such as BCE. This is why we need a new loss function to modulate the loss contributions from negative samples.

\subsection{Segmented Harmonic Loss} \label{harmonic_loss}
We use segment $\mathcal{F}_{r}$ for the formulation of the new loss function. After segmentation, each segment 
will contain classes with tolerable variance between their frequencies. However, the positive examples of any class from $\mathcal{F}_{r}$ 
(\textit{positive segment}) may be orders of magnitude less than negative samples contributed by the classes present in the remaining 
$\mathcal{S} -1$ segments (\textit{negative segments}). At the same time, these negative examples are necessary for the learning process 
of $\mathcal{F}_{r}$'s model to discriminate between positive and negative samples. 

Segmented Harmonic (SH) loss allows us to train models of different segments on the entire dataset of $N$ samples by achieving the following 
goals of our loss function:

\begin{itemize}[itemindent=2em]
  \item[\textbf{\emph{Goal 1:}}] Balance the loss contributions from the negative samples during training by dynamically weighing their losses.
  \item[\textbf{\emph{Goal 2:}}] Penalize classification errors for negative samples by dynamically increasing the corresponding loss.
  \item[\textbf{\emph{Goal 3:}}] Penalize classification errors for harder-to-classify samples by dynamically increasing the corresponding loss.
\end{itemize}

Since we are dealing with multi-label classification, a Binary Cross Entropy (BCE) loss provides a good starting point for the formulation. 
For simplicity, we shall denote the $k^{th}$ sample, so far denoted as $(\vect{x}^{k} ,\vect{y}^{k,r})$, as $(\vect{x}, \vect{y^r})$ removing the 
superscript $k$ indicating the sample number. Let $\vect{p^r}$ be the prediction of $\vect{y^r}$ and $\vect{y}$ be the original label before 
segmentation.  BCE Loss for a single class in $\mathcal{F}_{r}$ is given by:

\begin{equation}
  BCE(p^r, y^r) =
  \begin{cases}
    -log(p^r) & \text{if $y^r = 1$} \\
    -log(1 - p^r)  & \text{otherwise}
  \end{cases} \label{eq:bce}
\end{equation}

where $p^r \in [0, 1]$ is the prediction probability for the label $y^r=1$ in the single-class case. For notational convenience, we define $q^r$ as:
\begin{equation}
  q^r =
  \begin{cases}
    p^r & \text{if $y^r = 1$} \\
    1 - p^r  & \text{otherwise}
  \end{cases} \label{eq:qr}
\end{equation}
Using Equation \ref{eq:qr}, we rewrite BCE as:
\begin{equation}
  BCE(p^r, y^r) = BCE(q^r) = -log(q^r)
  \label{eq:bce_qr}
\end{equation}
Considering the multi-label scenario of segment $\mathcal{F}_{r}$, we can write the Cross Entropy (CE) loss using Equation \ref{eq:bce_qr} as:
\begin{equation}
  CE(\vect{q}^r)  = -\sum_{i=1}^{c_r}log({q}^r_{i})
  \label{eq:ce_qr}
\end{equation}
where:
\begin{conditions*}
	c_{r}				& :	& number of classes in segment $\mathcal{F}_{r}$, \\
	\vect{q}^r		& =	& $[{q}^r_{\rho + 1}, {q}^r_{\rho + 2}, ..., {q}^r_{\rho + c_r}]^T$, \\ 
	\rho				& : 	& total number of classes in the first $r - 1$ segments = $\sum_{p=1}^{r-1} c_{p}$, \\
\end{conditions*}

To meet \textbf{\textit{Goal 1}}, we need to include a modulating term in Equation \ref{eq:ce_qr} that will temper the contributions of the samples 
from negative segments which are far more in number than the positive examples. We start by calculating the approximate rates of occurrences of 
samples from the $i^{th}$ negative segment $\mathcal{F}_{i}$, w.r.t the positive samples from $\mathcal{F}_{r}$. 
Let $\mathbb{N}_i$ be the number of positive samples from segment $\mathcal{F}_{i}$. 
We define the approximate rate of occurrence of samples from $\mathcal{F}_{i}$ w.r.t $\mathcal{F}_{r}$ as:
\begin{equation}
  \mathbb{\beta}^{(i, r)}  = \frac{\mathbb{N}_i }{\mathbb{N}_r}
  \label{eq:beta_ir}
\end{equation}

If the label $\vect{y}^r$ is a negative sample i.e. a zero vector, its original label $\vect{y}$ must contribute to at least one segment other than 
$\mathcal{F}_{r}$. For example, if $\vect{y}$ contains two positive classes belonging to segment $\mathcal{F}_{l}$, and one to segment $\mathcal{F}_{m}$, 
we can approximate the mean rate for observing this particular negative sample $\vect{y}^r$ as:

\begin{equation}
  \beta^{SH} = \vect{mean}(\mathbb{\beta}^{(l, r)}, \mathbb{\beta}^{(l, r)}, \mathbb{\beta}^{(m, r)}) = \vect{mean}(2\mathbb{\beta}^{(l, r)}, \mathbb{\beta}^{(m, r)})
  \label{eq:beta_mean_eg}
\end{equation}

Different aggregate measures may be used to implement the $\vect{mean()}$ of Equation \ref{eq:beta_mean_eg}. Since we are aggregating rates, it is best implemented  
using harmonic mean.  Thus, we can rewrite Equation \ref{eq:beta_mean_eg} as:

\begin{equation}
  \beta^{SH} = \frac{ 2 + 1}{\frac{2}{\beta^{(l,r)}} + \frac{1}{\beta^{(m,r)}} }
  \label{eq:beta_mean_eg_hm}
\end{equation}

In general, if the original label $\vect{y}$ contain $n_i$ positive classes belonging to segment $\mathcal{F}_{i}$ where 
$ i \in \{ 1, 2, ..., \mathcal{S}\}$ and $ i \neq r$,the approximate the mean rate for observing the negative sample $\vect{y}^r$ is given by:

\begin{equation}
  \beta^{SH}_r = \frac{ \sum_{i=1}^\mathcal{S} n_i }{ \sum_{i=1}^\mathcal{S} \frac{n_i}{\beta^{(i,r)}} }
  \label{eq:beta_SH}
\end{equation}

In Equation \ref{eq:beta_SH}, since $\vect{y}^r$ is a negative sample originating from $\vect{y}$, there are no positive classes in $\vect{y}^r$, i.e. 
$n_r = 0$. Note that $\beta^{SH}_r $ can never be undefined which is discussed in Appendix \ref{secA3}. The Segmented Harmonic (SH) loss is given 
by the following equation by incorporating a modulating factor of $\frac{1}{\beta^{SH}_r}$:

\begin{equation}
  SH(\vect{q}^r)  = -\frac{1}{\beta^{SH}_r}\sum_{i=1}^{c_r}	log({q}^r_{i})
  \label{eq:sh_loss}
\end{equation}

Since $\beta^{SH}_r$ is the mean rate for observing the negative sample $\vect{y}^r$, a modulating factor inversely proportional to its occurrence rate will 
balance the contribution of this negative sample to the training loss, hence satisfying \textbf{\textit{Goal 1}}. We have found that introducing $\beta^{SH}_r$ 
in positive examples improves stability of the loss function. It is calculated in the same way as the negative examples. 

To meet the requirements of \textbf{\emph{Goal 2}} and \textbf{\emph{3}} we take inspiration from Focal Loss \cite{focal} and incorporate a term $ {(1- q^r)}^{\gamma}$ 
in Equation \ref{eq:sh_loss}. Thus we obtain SH Focal Loss as:

\begin{equation}
  SH_{Focal}(\vect{q}^r)  = -\frac{1}{\beta^{SH}_r}\sum_{i=1}^{c_r}{(1- q^r)}^{\gamma}log({q}^r_{i})
  \label{eq:sh_focal_qr}
\end{equation}

The focusing parameter $\gamma \geq 0$ from Focal Loss \cite{focal} penalizes the loss function for making mistakes on ``hard-to-classify'' instances. 
In our experiments, we tried $\gamma \in \{1.5, 2.0, 2.5 \}$ and found $2$, as suggested by the authors to work best. Details of Focal loss can be found in 
the paper \cite{focal}. 

\section{Experiments} \label{experiments}
We describe our experimental setup in Subsection \ref{ex_1}, its challenges and workarounds in Subsection \ref{ex_2}, and model selection in Subsection \ref{ex_3}.

\subsection{Experimental Setup} \label{ex_1}
Traditionally, the new era of machine learning has been widely successful because of GPU computing.  From safety-critical applications such as autonomous driving, \cite{gpu-av} to drug discovery \cite{gpu-drug},  GPU computing has been an essential ingredient, and much of this market (over $70 \%$) is dominated by NVIDIA \cite{nvidia-gpu}.  Almost all,  if not all, LLMs depend on NVIDIA's CUDA-enabled GPU architecture.  For initial experimental setup, coding, and debugging, we used a machine with a single NVIDIA A10G GPU with 24 GB of GPU Memory.  For training, we used a machine with four NVIDIA A10G GPUs, each GPU having a dedicated memory of 16 GB.  

\subsection{Setup Challanges} \label{ex_2}
Initially, we faced several implementation / resource-related challenges well-known to the machine-learning community, which mirrored the GPU resource-related problems faced by Deep Learning in its early days.  This problem is exacerbated in LLMs because of their sheer size. While the larger Deep Neural Networks had an average of 50 million parameters, even the simplest LLM, such as the base clinicalBERT, has a parameter count of 110 million.  GPT-J has 6 billion parameters, while the Llama-2 models start at 7 billion and go up to 70 billion parameters.  When trained without optimization, GPT-J requires a minimum of 90 GB to load.  To combat this resource-hungry aspect of LLMs, an entire sub-discipline of machine learning dedicated to optimizing GPU resources for the training and inference of LLMs has emerged \cite{gpu-research}.  The frameworks, libraries, and techniques we used to run these massive models are: 
\vspace{2pt}

\begin{enumerate}[label=\roman*.]
  \item \textbf{LoRA}: Low-Rank Adaptation, or LoRA \cite{lora}, is a framework that significantly reduces the trainable parameters by freezing the pretrained model weights and injecting trainable rank decomposition matrices into the LLM architecture \cite{lora}.  LoRA, known to reduce the number of trainable parameters by as much as 10,000 times and the GPU memory requirement by three times, was quite effective in running large LLMs such as Llama-2 on our smaller machines. 
  
  \item \textbf{Mixed-precision training}: This refers to the use of lower-precision formats than $float32$  such as $float16$ in an LLM during training, making it faster and use less memory while ensuring that compared to full precision training no task-specific accuracy is lost \cite{precision}.
  
  \item \textbf{Quantization}: It is the process of running inference by representing the weights and activations with lower precision data types, reducing memory footprint, such as replacing $float32$  with $float16$, making it possible to achieve as much as 2x reduction in memory usage \cite{quantization}.

  \item \textbf{xTuring}: xTuring \cite{xTuring} is an open-source library that simplifies the process of building, controlling, and tuning LLMs with a simple interface for personalizing the models to suit the application requirements.  This library includes many of the above optimizations, such as LoRA, and was instrumental in getting Llama-2 and GPT-J running. 
\end{enumerate}

\subsection{LLM Base Model Selection} \label{ex_3}
After initial experimentation on various architectures, we settled on a BERT encoder architecture for our base model over generative models such as GPT-J \cite{gptj} or Llama-2 \cite{llama2}.  A heavily quantized $int4$ version of GPT-J further optimized by LoRA (training only $1\%$ of its trainable parameters) fitted in our training machine's 16 GB GPU RAM.  Although it did not take significant time to run (~3 hours), the model did not converge and produced abysmal results.  Furthermore, being a generative model, the ICD-9 codes needed to be introduced in the vocabulary for the tuned model to predict the codes given an input note.  

We also performed experiments with Llama-2 (7B) using 8-bit quantization and optimized by LoRA (training $< 0.5 \%$ of its trainable parameters).  However, the time required to complete one epoch on the single GPU machine was ten days and 2.5 days on the training machine.  It was prohibitively expensive, and we discontinued the experiments after two days on the single GPU server. 

The BERT architecture fitted in our memory without mixed precision, but we used mixed precision for faster convergence.  Its bidirectional encoding was ideal for classification purposes, and unlike generative models, it did not require any addition to vocabulary as it treated the codes as classes.  Since we needed a deeper understanding of medical jargon and vocabulary, we settled on ClinicalBERT, a model identical in size and architecture to BERT but was imbued with medical vocabulary.  We also used it for data preprocessing.  ClinicalBERT was initialized from BERT and trained on a large dataset of EHRs from over 3 million patient records.  It performed much better (more than ten percentage points on micro F1 score) than Microsoft's PubMedBERT \cite{pubmedbert}, which was pretrained from scratch using abstracts from PubMed \cite{pubmed} as well as full-text articles from PubMedCentral.  The poor performance of PubMedBERT on the MIMIC dataset exemplifies the necessity of training base models on real-life data, which in our case are free-form handwritten notes and not fully formed text obtained from paper abstracts. 

\section{Results}
We trained our model for 150,000 steps or 8 hours, each step taking approximately $0.19$ seconds. Using stratified sampling, we split the dataset into training, validation and test datsets in the ratio of $94:3:3$. Stratified sampling ensured that our split was unbiased, and captured key population characteristics i.e. included all classes including minority classes as in the training set. 

\begin{table*}[h]
\centering
\caption{Micro F1 scores for the total MIMIC Dataset and its different segments}\label{tab1}%
\begin{tabular}{@{}llllllll@{}}
\toprule
\textbf{Methods}		& Total 		& Head  	& Body 1 	& Body 2 	& Body 3 	& Body 4 	& Tail\\
\midrule
BCE-BERT    		 	& 69.87		& 70.74		& 68.61		& 65.40		& 58.96		& 54.31		& 46.03 \\
Focal-BERT    			& \textbf{71.42}		& 72.97   	& 73.83  	& 68.93  	& 63.00 	& 60.23  	& 42.50 \\
CB-Focal-BERT    		& 69.91   	& 71.01  	& 71.93  	& 67.93 	& 61.07 	& 58.42 	& 51.87\\
SH-Focal-BERT    		& 71.28   	& \textbf{72.73}  	& \textbf{77.01}  	& \textbf{71.31}  	& \textbf{64.22} 	& \textbf{61.63} & \textbf{59.81}\\
\botrule
\end{tabular}
\end{table*}

For all experiments, we used AdamW optimizer \cite{adamw} with default settings except for the learning rate, which we initialized to $0.0005$.  We used Binary Cross Entropy (BCE) as the baseline loss function.  Since our data is highly imbalanced, we did not try resampling techniques \cite{resampling} because they are not effective for multi-label scenarios in NLP.  Instead, we tried various loss functions such as Focal Loss \cite{focal} and Class-Balanced Loss (CB) \cite{CBLoss}, which are designed to mitigate class imbalance in multi-label scenarios.  For Focal Loss, the focusing hyperparameter $\gamma$ that worked best for us is 2.  For Class-Balanced Loss, we set the hyperparameter $\beta$ to 0.99.  The $\beta^{SH}_r$ for Segmented Harmonic Loss is calculated based on the dataset and the hyperparameter $\eta_{\mathcal{S}}$, which we set to a value of 0.5.

\begin{figure*}[h]
\centering
\includegraphics[width=1.0\textwidth]{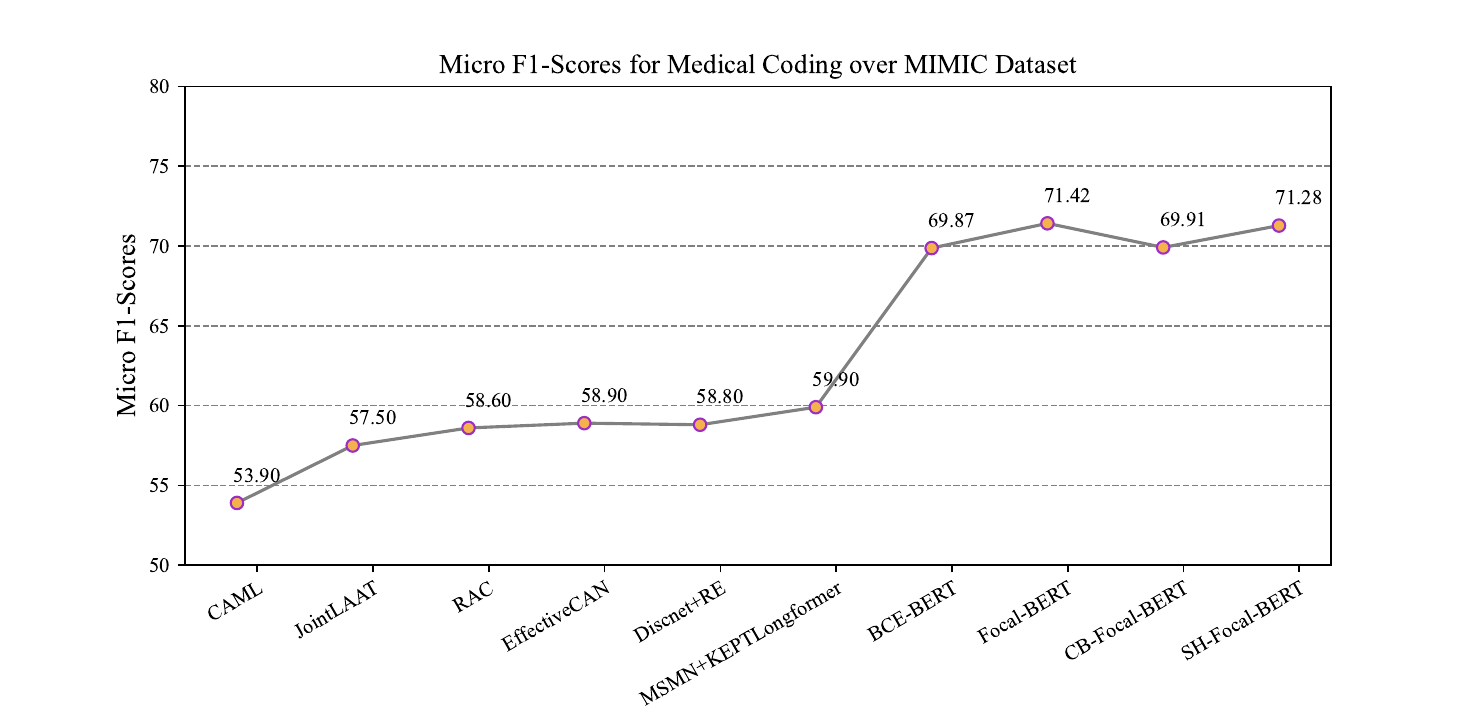}
\caption{Mirco F1 score comparisons with well known approaches: CAML \cite{caml}, JointLAAT \cite{LAAT}, RAC\cite{RAC}, EffectiveCAN\cite{can}, Discnet+RE\cite{discnetre}, MSMN+LongFormer\cite{longformer}. }\label{figlast}
\end{figure*}

For performance metrics, we used \textit{segment-wise micro F1 score}.  Since macro F1 scores treat all classes equally regardless of their support values,  a simple average may be a misleading performance metric for imbalanced distributions.  Micro F1 score, which gives equal weights to each sample regardless of its class, is also not particularly suitable as a majority class can dominate its value.  Instead, we segmented the imbalanced distribution using the equation \ref{eq:sigma}.  The hyperparameter $\eta_{\mathcal{S}}$'s value segmented the dataset into six segments with comparable variance.  For each segment, we calculated a micro F1 score or \textit{segment-wise micro F1 score}.

Table \ref{tab1} shows the performance of different losses with clinicalBERT as the base.  Segmented Harmonic Focal Loss (abbreviated as SH-Focal-BERT) performs much more evenly than Focal Loss (Focal-BERT), whose micro F1 scores degrade towards the tail end.  The same happens for Class-Balanced Focal loss (CB-Focal-BERT), though it fares better towards the tail region.  Segmented Harmonic Focal Loss performs the best through all segments, though its total micro F1 score is marginally less than the Focal-BERT configuration.

In general, the effect of LLMs over other approaches is evident from Figure \ref{figlast}.  There is a ten percentage point jump from the approach in \cite{longformer}.  Although most of these approaches worked with only the MIMIC III dataset and did not enhance their dataset with records from MIMIC IV, the performance gain is not solely due to the larger dataset.  Our initial experiments with only the MIMIC III dataset showed a seven percentage point gain over previous approaches such as in \cite{longformer}.

\section{Conclusion \& Future Work}\label{sec13}
The fact that LLMs would outperform previous architectures by as much as ten percentage points was not surprising to us.  Vaswani et al.'s transformer architecture has proven to be superior over Recurrent Neural Networks over and over again and has rewritten the book in NLP.  With Harmonic Loss developed to tackle multi-label extreme imbalance, the results improved even more for the minority classes.  It would not be surprising either if a different approach to cleaning and dealing with class imbalance yields even better results.  

However, our experiments, particularly our failure to tackle the extra-large models, reveal something more telling: domain-specific training of LLMs can be prohibitively expensive until the hardware catches up with the rapid progress in R\&D.  The healthcare sector has always been wary of using open-source software.  LLMs trained on a big chunk of the internet, books, and other text sources are subject to litigations \cite{litigation}, and healthcare companies are wary of it.  It hinders the adoption of this marvel of technology.   

Another key takeaway is a bigger LLM is not essential for good results.  A highly regulated domain like healthcare requires quality control and replicability of results.  In our future work, those are the aspects we would like to explore:  how do we build sufficiently large LLMs from scratch in-house, one that protects patients' privacy, have stringent quality control measures, and replicability of results?  We believe that would unlock the full potential of LLMs in healthcare.

\begin{appendices}

\section{Pseudocode for procedure \textsc{SegmentTail}}\label{secA2}
A Binary-Search-based implementation of the procedure \textsc{SegmentTail} is given below. \textsc{Std} is any procedure to
compute standard deviation of a list.
\begin{algorithm}
\caption{Return the starting index of the tail segment $\mathcal{F}_{\mathcal{S}}$ }
\begin{algorithmic}[1]
\Procedure{SegmentTail}{$\mathcal{F}$, $\eta_{\mathcal{S}}$}
	\State \text{//$\mathcal{F}$: index-sorted list of class frequencies}
	\State $n \gets$ {$len(\mathcal{F})$}
	\State \text{// Initialize the left pointer for Binary Search }
	\State $left \gets 0$ 
	\State \text{// Initialize the right pointer for Binary Search }
	\State $right \gets (n - 1)$ 
	\State \text{// Compute the allowed standard deviation} using $\eta_{\mathcal{S}}$ and the last frequency of $\mathcal{F}$
	\State $\sigma_{allowed} \gets \eta_{\mathcal{S}} \times \mathcal{F}[n - 1]$
	
	\State \text{// Find the starting index of the tail segment $\mathcal{F}_{\mathcal{S}}$}
	\While { $left <  right$} 
	\State \text{// Get the middle index }
	\State $mid \gets (left + right)/2$
	\State \text{// Get the standard deviation of $\mathcal{F}[mid:]$ }
	\State $\sigma_{mid} \gets$ \Call{Std}{$\mathcal{F}[mid:]$}
	\If{$\sigma_{mid} > \sigma_{allowed}$}
    \State $left = mid + 1$
  	\ElsIf{$\sigma_{mid} < \sigma_{allowed}$}
    \State $right = mid$
  	\EndIf	
	\EndWhile \\
    \Return $left$
\EndProcedure
\end{algorithmic}
\label{alg:segment_once}
\end{algorithm}

\section{Differentiablity of Segmented Harmonic Loss}\label{secA3}
For $\vect{q}^r \in (0, 1)$, Segmented Harmonic Loss as defined in equation \ref{eq:sh_focal_qr} is differentiable.
Any negative sample $\vect{y}^r$ belonging to $\mathcal{F}_{r}$ must contribute at least one positive sample belonging to other
segments. This will result in a positive harmonic mean $\beta^{SH}_r$. Thus, the modulating term $\frac{1}{\beta^{SH}_r}$  
is always positive and can never be undefined. The remaining factor in equation \ref{eq:sh_focal_qr} is essentially BCE 
and Focal Loss whose derivatives and differentiablity has been proven in \cite{focal}.




\end{appendices}

\bibliographystyle{plain}
\bibliography{cmx}

\end{document}